\def\year{2018}\relax
\documentclass[letterpaper]{article} 
\usepackage{aaai18}  
\usepackage{times}  
\usepackage{helvet}  
\usepackage{courier}  
\usepackage{url}  
\usepackage{graphicx}  
\newcommand{\newcite}[1]{\citeauthor{#1}~(\citeyear{#1})}

\usepackage{amsmath,mathrsfs,bbm}
\usepackage{booktabs}

\usepackage{multirow}
\usepackage{amssymb,bm}
\DeclareMathOperator*{\argmax}{argmax}
\usepackage{graphicx}
\usepackage{subfigure}
\usepackage{enumitem}
\usepackage{multirow}
\usepackage{url}

\begin{document}
%
\title{Order-Planning Neural Text Generation From Structured Data}

\author{Lei Sha,\!\!\,$^1$ Lili Mou,\!\!\,$^2$ Tianyu Liu,\!\!\,$^1$ Pascal Poupart,\!\!\,$^2$ Sujian Li,\!\!\,$^1$ Baobao Chang,\!\!\,$^1$ Zhifang Sui$^1$\\
	$^1$Key Laboratory of Computational Linguistics, Ministry of Education; School of EECS, Peking Univeristy\\
	$^2$David R. Cheriton School of Computer Science, University of Waterloo\\
	$^1$\{shalei, tianyu0421, lisujian, chbb, szf\}@pku.edu.cn\\
	$^2$doublepower.mou@gmail.com, ppoupart@uwaterloo.ca
}
\maketitle

\begin{abstract}
Generating texts from structured data (e.g., a table) is important for various natural language processing tasks such as question answering and dialog systems. In recent studies, researchers use neural language models and encoder-decoder frameworks for table-to-text generation. However, these neural network-based approaches do not model the order of contents during text generation. When a human writes a summary based on a given table, he or she would probably consider the content order before wording. In a biography, for example, the nationality of a person is typically mentioned before occupation in a biography. In this paper, we propose an order-planning text generation model to capture the relationship between different fields and use such relationship to make the generated text more fluent and smooth. We conducted experiments on the \textsc{WikiBio} dataset and achieve significantly higher performance than previous methods in terms of BLEU, ROUGE, and NIST scores.
\end{abstract}

\section{Introduction}

Generating texts from structured data (e.g., a table) is important for various natural language processing tasks such as question answering and dialog systems. Table~\ref{tab:example} shows an example of a Wikipedia infobox (containing fields and values) and a text summary.

In early years, text generation is usually accomplished by human-designed rules and templates~\cite{green2006generation,turner2010generating}, and hence the generated texts are not flexible. Recently, researchers apply neural networks to generate texts from structured data~\cite{wikibio}, where a neural encoder captures table information and a recurrent neural network (RNN) decodes these information to a natural language sentence.

\begin{table}[!t]
	\textbf{Table:}\\[.2cm]
	\resizebox{\linewidth}{!}{
	\footnotesize
	\begin{tabular}{r|ll|}
		\cline{2-3}
		\textbf{ID}\!\! &\!\! \textbf{Field} & \textbf{Content}\\
		\cline{2-3}
		1\!\! &\!\! Name  \!\!&\!\! \textit{Arthur Ignatius Conan Doyle}\\
		2\!\! &\!\! Born  \!\!&\!\! \textit{22 May 1859 Edinburgh, Scotland}\\
		3\!\! &\!\! Died  \!\!&\!\! \textit{7 July 1930 (aged 71) Crowborough, England}\!\!\\
		4\!\! &\!\! Occupation \!\!&\!\! \textit{Author, writer, physician}\\
		5\!\! &\!\! Nationality \!\!&\!\! \textit{British}\\
		6\!\! &\!\! Alma mater \!\!&\!\! \textit{University of Edinburgh Medical School}\\
		7\!\! &\!\! Genre \!\!&\!\! \textit{Detective fiction fantasy}\\
		8\!\! &\!\! Notable work \!\!\!&\!\! \textit{Stories of Sherlock Homes}\\
		\cline{2-3}
	\end{tabular}
}\\[.2cm]

	\textbf{Text:}\\[.1cm]
	{\small\verb|  |Sir Arthur Ignatius Conan Doyle (22 May 1859 -- 7 July 1930) was a British writer best known for his detective fiction featuring the character Sherlock Holmes.}
	
	\caption{Example of a Wikipedia infobox and a reference text.}\label{tab:example}
\end{table}

Although such neural network-based approach is capable of capturing complicated language and can be trained in an end-to-end fashion, it lacks explicit modeling of content order during text generation. That is to say, an RNN generates a word at a time step conditioned on previous generated words as well as table information, which is more or less ``shortsighted'' and differs from how a human writer does. As suggested in the book \textit{The Elements of Style},
\begin{quote}
	A basic structural design underlies every kind of writing \dots\
	in most cases, planning must be a deliberate prelude to writing. \cite{element}
\end{quote}
This motivates order planning for neural text generation. In other words, a neural network should model not only word order (as has been well captured by RNN) but also the order of contents, i.e., fields in a table.

We also observe from real summaries that table fields by themselves provide illuminating clues and constraints of text generation. In the biography domain, for example, the nationality of a person is typically mentioned before the occupation. This could benefit from explicit planning of content order during neural text generation.

In this paper, we propose an order-planning method for table-to-text generation. Our model is built upon the encoder-decoder framework and use RNN for text synthesis with attention to table entries. Different from exiting neural models, we design a table field linking mechanism, inspired by temporal memory linkage in the Differentiable Neural Computer~\cite[DNC]{DNC}. Our field linking mechanism explicitly models the relationship between different fields, enabling our neural network to better plan what to say first and what next. Further, we  incorporate a copy mechanism~\cite{copynet} into our model to cope with rare words.

We evaluated our method on the \textsc{WikiBio} dataset~\cite{wikibio}. Experimental results show that our order-planning approach significantly outperforms previous state-of-the-art results in terms of BLEU, ROUGE, and NIST metrics. Extensive ablation tests verify the effectiveness of each component in our model; we also perform visualization analysis to  better understand the proposed order-planning mechanism.

\section{Approach}

Our model takes as input a table (e.g., a Wikipedia infobox) and generates a natural language summary describing the information based on an RNN. The neural network contains three main components:
\begin{itemize}
	\item An encoder captures table information;
	\item A dispatcher---a hybrid content- and linkage-based attention mechanism over table contents---plans what to generate next; and
	\item A decoder generates a natural language summary using RNN, where we also incorporate a copy mechanism~\cite{copynet} to cope with rare words.
\end{itemize}
We elaborate these components in the rest of this section.

\subsection{Encoder: Table Representation}

\begin{figure}[!t]
	\begin{center}
		\begin{tabular}{ccc}
		 &\textbf{(a) Encoder} & \textbf{(b) Dispatcher}\\
			&\footnotesize	Table Representation &\footnotesize Planning What to Generate Next
		\end{tabular}
		\includegraphics[width=\linewidth]{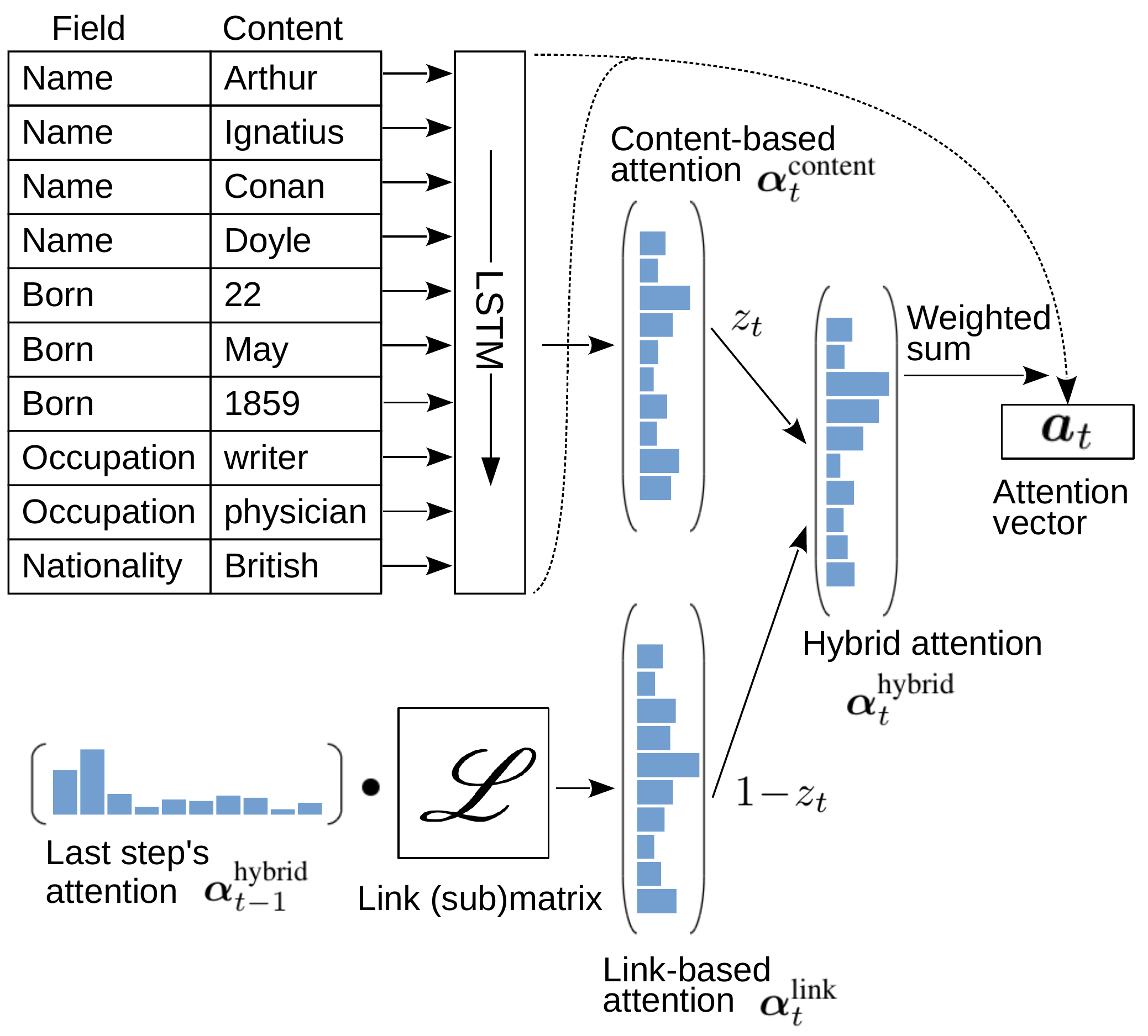}
		\caption{The (a) Encoder and (b) Dispatcher in our model.}
		\label{fig:arch}
	\end{center}
\end{figure}

We design a neural encoder to represent table information.
As shown in Figure~\ref{fig:arch}, the content of each field is split into separate words and the entire table is transformed into a large sequence. Then we use a recurrent neural network (RNN) with long short term memory (LSTM) units \cite{lstm} to read the contents as well as their corresponding field names.

Concretely, let $C$ be the number of content words in a table; let $\bm c_i$ and $\bm f_i$ be the embeddings of a content and its corresponding field, respectively ($i=1\cdots C$). The input of LSTM-RNN is the concatenation of $\bm f_i$ and $\bm c_i$, denoted as $\bm x_i=[\bm f_i; \bm c_i]$, and the output, denoted as $\bm h_i$, is the encoded information corresponding to a content word, i.e.,
\begin{align}\label{eq:lstm:begin}
\big[\bm g_\text{in}; \bm g_\text{forget}; \bm g_\text{out}\big]  &= \sigma(W_g\bm x_i+U_g\bm h_{i-1}) \\
\widetilde{\bm x}_i&=\tanh(W_x\bm x_i+U_x\bm h_{i-1}) \\
\widetilde{\bm h}_i &= \bm g_\text{in}\circ \widetilde{\bm x}_i + \bm g_\text{forget}\circ \widetilde{\bm h}_{i-1}\\
\bm h_i &= \bm g_\text{out} \circ \tanh(\widetilde{\bm h}_i)\label{eq:lstm:end}
\end{align}
where $\circ$ denotes element-wise product, and $\sigma$ denotes the $\operatorname{sigmoid}$ function. $W$'s and $U$'s are weights, and bias terms are omitted in the equations for clarity. $\bm g_\text{in}$, $\bm g_\text{forget}$, and $\bm g_\text{out}$ are known as input, forget, and output gates.

Notice that, we have two separate embedding matrices for fields and content words. We observe the field names of different data samples  mostly come from a fixed set of candidates, which is reasonable in a particular domain. Therefore, we assign an embedding to a field, regardless of the number of words in the field name. For example, the field \textit{Notable work} in Table~\ref{tab:example} is represented by a single field embedding instead of the embeddings of \textit{notable} and \textit{work}.

For content words, we represent them with conventional word embeddings (which are randomly initialized), and use LSTM-RNN to integrate information. In a table, some fields contain a sequence of words (e.g., \textit{Name}=``\textit{Arthur Ignatius Conan Doyle}''), whereas other fields contain a set of words (e.g., \textit{Occupation} = ``\textit{writer, physician}''). We do not have much human engineering here, but let an RNN to capture such subtlety by itself.

\subsection{Dispatcher: Planning What to Generate Next}

After encoding table information, we use another RNN to decode a natural language summary (deferred to the next part). During the decoding process, the RNN is augmented with a dispatcher that plans what to generate next.

Generally, a dispatcher is an attention mechanism over table contents. At each decoding time step $t$, the dispatcher computes a probabilistic distribution $\alpha_{t,i}$ ($i=1\cdots C$), which is further used for weighting content representations~$\bm h_i$.
In our model, the dispatcher is a hybrid of content- and link-based attention, discussed in detail as follows.

\subsubsection{Content-Based Attention.} Traditionally, the computation of attention $\alpha_{t,i}$ is based on the content representation $\bm h_i$ as well as some state during decoding \cite{attention,mei}. We call this \textit{content-based attention}, which is also one component in our dispatcher.

Since both the field name and the content contain important clues for text generation, we compute the attention weights based on not only the encoded vector of table content  $\bm h_{i}$ but also the field embedding $\bm f_i$, thus obtaining the final attention $\alpha_{t,i}^\text{content}$ by re-weighting one with the other. Formally, we have
\begin{align}\label{eqn:content1}
\widetilde{\alpha}_{t,i}^{(f)} &= \bm f_i^\top\big(W^{(f)}\bm y_{t-1}+\bm b^{(f)}\big)\\\label{eqn:content2}
\widetilde{\alpha}_{t,i}^{(c)} &=\bm h_i^\top\big(W^{(c)}\bm y_{t-1}+\bm b^{(c)}\big)\\\label{eqn:content}
\alpha_{t,i}^\text{content}&=\dfrac{\exp\big\{\widetilde{\alpha}_{t,i}^{(f)}\widetilde{\alpha}_{t,i}^{(c)}\big\}}
{\sum_{j=1}^C\exp\big\{\widetilde{\alpha}_{t,j}^{(f)}\widetilde{\alpha}_{t,j}^{(c)}\big\}}
\end{align}
where $W^{(f)}, \bm b^{(f)}, W^{(c)}, \bm b^{(c)}$ are learnable parameters; $\bm f_i$ and $\bm h_i$ are vector representations of the field name and encoded content, respectively, for the $i$th row. $\alpha_{t,i}^\text{content}$ is the content-based attention weights. Ideally, a larger content-based attention indicates a more relevant content to the last generated word.

\subsubsection{Link-Based Attention.}
We further propose a \textit{link-based attention} mechanism that directly models the relationship between different fields.

Our intuition stems from the observation that, a well-organized text typically has a reasonable order of its contents. As illustrated previously, the nationality of a person is often mentioned before his occupation (e.g., \textit{a British writer}). Therefore, we propose an link-based attention to explicitly model such order information.

We construct a link matrix $\mathscr{L}\in\mathbb{R}^{n_{\!f}\times n_{\!f}}$, where $n_{f}$ is the number possible field names in the dataset. An element $\mathscr{L}[f_{\!j},f_{\!i}]$ is a real-valued score indicating how likely the field $f_j$ is mentioned after the field $f_i$.  (Here, $[\cdot, \cdot]$ indexes a matrix.) The link matrix $\mathscr{L}$ is a part of model parameters and learned by backpropagation. Although the link matrix appears to be large in size (1475$\times$1475), a large number of its elements are not used because most fields do not co-occur in at least one data sample; in total, we have 53422 effective parameters here. In other scenarios, low-rank approximation may be used to reduce the number of parameters.

Formally, let $\alpha_{t-1,i}$ ($i=1\dots C$) be an attention probability\footnote{Here, $\alpha_{t-1,i}$ refers to the hybrid content- and link-based attention, which will be introduced shortly.}  over table contents in the last time step during generation. For a particular data sample whose content words are of fields $f_1, f_2, \cdots, f_C$, we first weight the linking scores by the previous attention probability, and then normalize the weighted score to obtain link-based attention probability, given by
\begin{align}
\alpha_{t,i}^\text{link}&=\text{softmax}\bigg\{\sum_{j=1}^C\alpha_{t-1,j}\cdot\mathscr{L}[f_{\!j},f_{\!i}]\bigg\}\\
&=\dfrac{\exp\big\{
		\sum_{j=1}^C\alpha_{t-1,j}\cdot\mathscr{L}[f_{\!j},f_{\!i'}]
	\big\}
	}{\sum_{i'=1}^C\exp\big\{\sum_{j}\alpha_{t-1,j}\cdot\mathscr{L}[f_{\!j},f_{\!i'}]\big\}}\label{eq:link}
\end{align}

Intuitively, the link matrix is analogous to the transition matrix in a Markov chain~\cite{stochasticprocess}, whereas the term $\sum_{j=1}^C\alpha_{t-1,j}\cdot\mathscr{L}[f_{\!j},f_{\!i}]$ is similar to one step of transition in the Markov chain. However, in our scenario, a table in a particular data sample contains only a few fields, but a field may occur several times because it contains more than one content words. Therefore, we do not require our link matrix $\mathscr{L}$ to be a probabilistic distribution in each row, but normalize the probability afterwards by Equation~\ref{eq:link}, which turns out to work well empirically.

Besides, we would like to point out that the link-based attention is inspired by the Differentiable Neural Computer~\cite[DNC]{DNC}. DNC contains a ``linkage-based addressing'' mechanism to track consecutively used memory slots and thus to integrate order information during memory addressing. Likewise, we design the link-based attention to capture the temporal order of different fields. But different from the linking strength heuristically defined in DNC, the link matrix in our model is directly parameterized and trained in an end-to-end manner.

\subsubsection{Hybrid Attention.}
To combine the above two attention mechanisms, we use a self-adaptive gate $z_t\in(0,1)$ by a sigmoid unit
\begin{align}
z_t &= \sigma\big(\bm w^\top[\bm h'_{t-1}; \bm e_t^{(f)}; \bm y_{t-1} ]\big)
\end{align}
where $\bm w$ is a parameter vector. $\bm h'_{t-1}$ is the last step's hidden state of the decoder RNN. $\bm y_{t-1}$ is the embedding of the word generated in the last step; $\bm e_t^{(f)}$ is the sum of field embeddings $\bm f_i$ weighted by the current step's field attention $\alpha_{t,i}^\text{link}$. As $\bm y_{t-1}$ and $\bm e_t^{(f)}$ emphasize the content and link aspects, respectively, the self-adaptive gate $z$ is aware of both.
In practice, we find $z$ tends to address link-based attention too much and thus adjust it by $\widetilde z_t= 0.2z_t + 0.5$ empirically.

Finally, the hybrid attention, a probabilistic distribution over all content words, is given by
\begin{equation}
 \bm\alpha_t^\text{hybrid} = \widetilde z_t \cdot \bm \alpha_t^\text{content} + (1-\widetilde z_t)\cdot \bm \alpha_t^\text{link}\label{eq:gate}
\end{equation}

\subsection{Decoder: Sentence Generation}

\begin{figure}[!t]
	\begin{center}
		\textbf{Decoder}\\
		Sentence Generation
		
		\smallskip
		
		\includegraphics[width=.9\linewidth]{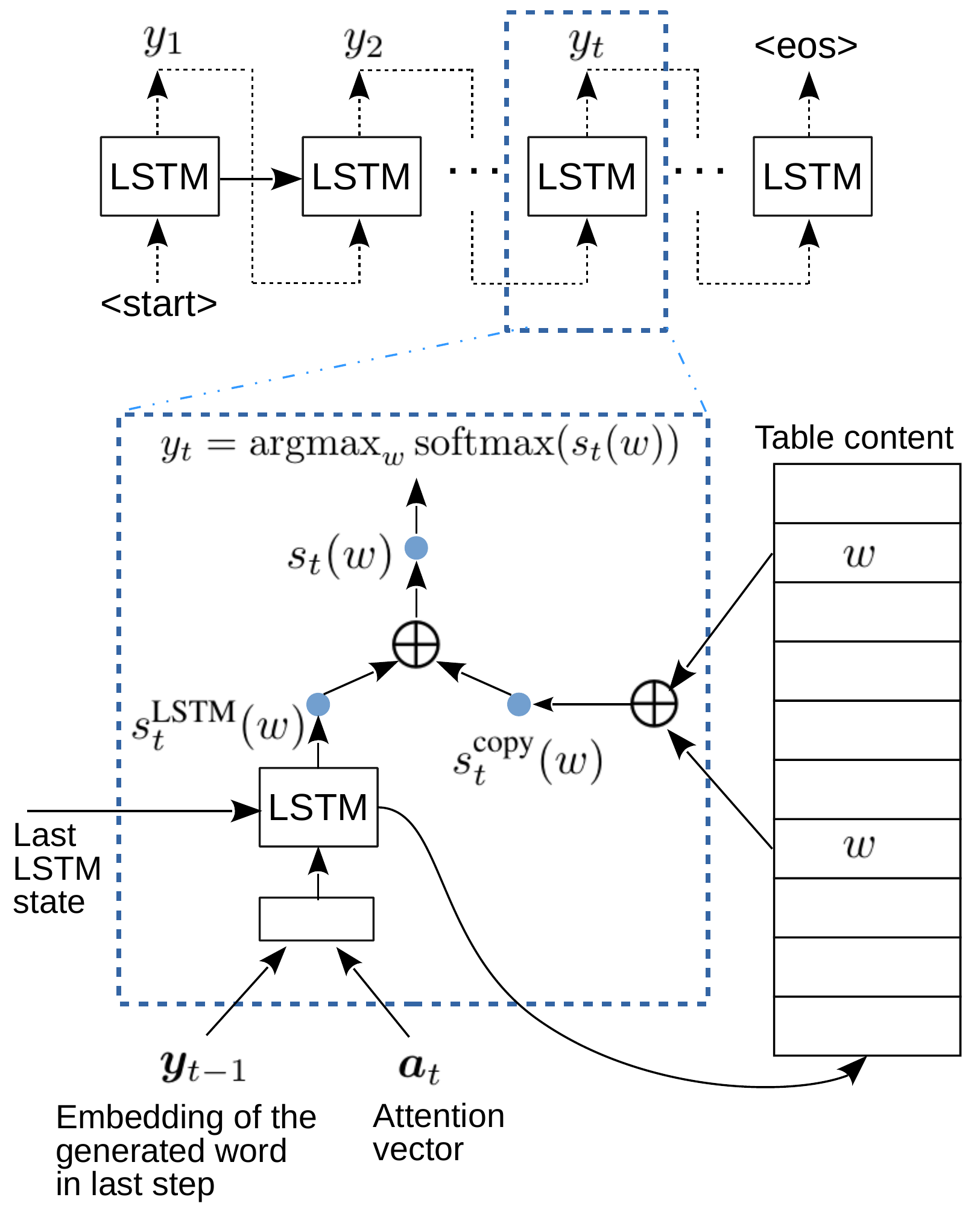}
		\caption{The decoder RNN in our model, which is enhanced with a copy mechanism.}
	\end{center}
\end{figure}

The decoder is an LSTM-RNN that predicts target words in sequence. We also have an attention mechanism \cite{attention} that summarizes source information, i.e., the table in our scenario, by weighted sum, yielding an attention vector $\bm a_t$ by
\begin{equation}
\bm a_t = \sum_{i=1}^C\alpha_{t,i}^\text{hybrid}\bm h_i
\end{equation}
where $\bm h_i$ is the hidden representation obtained by the table encoder. As $\alpha_{t,i}^\text{hybrid}$ is a probabilistic distribution---determined by both content and link information---over content words, it enables the decoder RNN to focus on relevant information at a time, serving as an order-planning mechanism for table-to-text generation.

Then we concatenate the attention vector $\bm a_t$ and the embedding of the last step's generated word $\bm y_{t-1}$, and use a single-layer neural network to mix information before feeding to the decoder RNN. In other words, the decoder RNN's input (denoted as $\bm x_t$) is
\begin{equation}
\bm x_t=\tanh(W_d[\bm a_t; \bm y_{t-1}]+\bm b_d)\label{eqn:RNNinput}
\end{equation}
where $W_d$ and $b_d$ are weights. Similar to Equations~\ref{eq:lstm:begin}--\ref{eq:lstm:end}, at a time step $t$ during decoding, the decoder RNN yields a hidden representation $\bm h_t'$, based on which a score function $\bm s^\text{LSTM}$ is computed suggesting the next word to generate. The score function is computed by
\begin{equation}
\bm s_t^\text{LSTM}=W_s\bm h_t'+\bm b_s
\end{equation}
where $\bm h_t'$ is the decoder RNN's state. ($W_s$ and $\bm  b_s$ are weights.) The score function can be thought of as the input of a softmax layer  for classification before being normalized to a probabilistic distribution. We incorporate a copy mechanism~\cite{copynet} into our approach, and the normalization is accomplished after considering  a copying score, introduced as follows.

\begin{table*}[!t]
	\begin{center}
		\begin{tabular}{llrrr}\toprule
			\textbf{Group}&\textbf{Model} & \textbf{BLEU} &\textbf{ROUGE} &\textbf{NIST}\\
			\midrule
			Previous results\ \ \ \  &	KN & 2.21& 0.38& 0.93\\
			&Template KN      &19.80 & 10.70& 5.19\\
			\cmidrule{2-5}
			&	Table NLM$^l$ & 34.70&25.80 &7.98 \\\midrule
			Our results & Content attention only\ \ \ \ &41.38 &34.65 & 8.57\\
			& Order planning (full model)         &\textbf{43.91} &\textbf{37.15} & \textbf{8.85}\\
			\bottomrule
		\end{tabular}
	\end{center}
	\vspace{-.2cm}
	\caption{Comparison of the overall performance between our model and previous methods. $^l$Best results in \protect\newcite{wikibio}.}
	\vspace{-.3cm}
	\label{tab:overall}
\end{table*}
\subsubsection{Copy Mechanism.}
The copy mechanism scores a content word $c_i$ by its hidden representation $\bm h_i$ in the encoder side, indicating how likely the content word $c_i$ is directly copied during target generation. That is,
\begin{equation}
s_{t,i}=\sigma(\bm h_i^\top W_c)\bm h_{t}'
\end{equation}
and $s_{t,i}$ is a real number for $i=1,\cdots, C$ (the number of content words). Here $W_c$ is a parameter matrix, and $\bm h'$ is the decoding state.

In other words, when a word appears in the table content, it has a copying score computed as above. If a word $w$ occurs multiple times in the table contents, the scores are added, given by
\begin{equation}
s_t^\text{copy}(w)= \sum_{i=1}^Cs_{t,i} \cdot \mathbbm{1}_{\{c_i=w\}}
\end{equation}
where $\mathbbm{1}_{\{c_i=w\}}$ is a Boolean variable indicating whether the content word $c_i$ is the same as the word $w$ we are considering.

Finally, the LSTM score and the copy score are added for a particular word and
further normalized to obtain a probabilistic distribution, given by
\begin{align}
s_t(w)&=s_t^\text{LSTM}(w) + s_t^\text{copy}(w)\\
p_t(w)&=\operatorname{softmax}\left(s_t(w)\right)=\frac{\exp\{s_t(w)\}}{\sum\limits_{w'\in\mathcal{V}\bigcup\mathcal{C}}\exp\{s_t(w')\}}
\label{eq:predict}
\end{align}
where $\mathcal{V}$ refers to the vocabulary list and $\mathcal{C}$ refers to the set of content words in a particular data sample.
In this way, the copy mechanism can either generate a word from the vocabulary or directly copy a word from the source side. This is hepful in our scenario because some fields in a table (e.g., \textit{Name}) may contain rare or unseen words and the copy mechanism can cope with them naturally.

For simplicity, we use greedy search during inference, i.e., for each time step $t$, the word with the largest probability is chosen, given by
$y_t=\argmax_w{p_t(w)}$. The decoding process terminates when a special symbol $<$eos$>$ is generated, indicating the end of a sequence.

\subsection{Training Objective}
Our training objective is the negative log-likelihood of a sentence $y_1\cdots y_T$ in the training set.
\begin{equation}\label{eq:obj}
J=-\sum_{t=1}^T\log p(y_t|y_0\cdots y_{t-1})
\end{equation}
where $p(y_t|\cdot)$ is computed by Equation~\ref{eq:predict}. An $\ell_2$ penalty is also added as most other studies.

Since all the components described above are differentiable, our entire model can be trained end-to-end by backpropagation, and we use Adam~\cite{adam} for optimization.
\section{Experiments}

\subsection{Dataset}

We used the newly published \textsc{WikiBio} dataset \cite{wikibio},\footnote{\url{https://github.com/DavidGrangier/wikipedia-biography-dataset}} which contains 728,321 biographies from WikiProject Biography\footnote{\url{https://en.wikipedia.org/wiki/Wikipedia:WikiProject_Biography}} (originally from English Wikipedia, September 2015).

Each data sample comprises an infobox table of field-content pairs, being the input of our system. The generation target is the first sentence in the biography, which follows the setting in previous work \cite{wikibio}. Although only the first sentence is considered in the experiment, the sentence typically serves as a summary of the article. In fact, the target sentence has 26.1 tokens on average, which is actually long. Also, the sentence contains information spanning multiple fields, and hence our order-planning mechanism is useful in this scenario.

We applied the standard data split: 80\% for training and 10\% for testing, except that model selection was performed on a validaton subset of 1000 samples (based on BLEU-4).

\subsection{Settings}

We decapitalized all words and kept a vocabulary size of 20,000 for content words and generation candidates, which also followed previous work~\cite{wikibio}. Even with this reasonably large vocabulary size, we had more than 900k out-of-vocabulary words. This rationalizes the use of the copy mechanism.

For the names of table fields, we treated each as a special token. By removing nonsensical fields whose content is ``none'' and grouping fields occurring less than 100 times as an ``Unknown'' field, we had 1475 different field names in total.

In our experiments, both words' and table fields' embeddings were 400-dimensional and LSTM layers were 500-dimensional. Notice that, a field (e.g., ``name'') and a content/generation word (e.g., also ``name''), even with the same string, were considered as different tokens; hence, they had different embeddings. We randomly initialized all embeddings, which are tuned during training.

We used Adam~\cite{adam} as the optimization algorithm with a batch size of 32; other hyperparameters were set to default values.

\subsection{Baselines}
We compared our model with previous results using either traditional language models or neural networks.
\begin{itemize}
	\item KN and Template KN~\cite{heafield2013scalable}: \newcite{wikibio} train an interpolated Kneser-Ney (KN) language model for comparison with the KenLM toolkit. They also train a KN language model with templates.
	
	\item Table NLM: \newcite{wikibio} propose an RNN-based model with attention and copy mechanisms. They have several model variants, and we quote the highest reported results.
\end{itemize}

We report model performance in terms of several metrics, namely BLEU-4, ROUGE-4, and NIST-4, which are computed by standard software, NIST mteval-v13a.pl (for BLEU and NIST) and MSR rouge-1.5.5 (for ROUGE). We did not include the perplexity measure in \newcite{wikibio} because the copy mechanism makes the vocabulary size vary among data samples, and thus the perplexity is not comparable among different approaches.

\subsection{Results}

\begin{table}[!t]
	\centering
	\begin{tabular}{lccc}
		\toprule
		\!\!\textbf{Component} & \textbf{BLEU} & \textbf{ROUGE} & \textbf{NIST}\\
		\midrule
		\!\!Content att. &41.38 &34.65 & 8.57\\
		\!\!Link att.   &38.24 &32.75 &8.36 \\
		\!\!Hybrid att. &43.01 &36.91 & 8.75\\
		\midrule
		\!\!Copy$+$Content att. &41.89 &34.93 & 8.63\\
		\!\!Copy$+$Link att.    &39.08 &33.47 &8.42 \\
		\!\!Copy$+$Hybrid att.  &\textbf{43.91} &\textbf{37.15} & \textbf{8.85}\\
		\bottomrule
	\end{tabular}
	\caption{Ablation test.}\label{tab:ablation}
\end{table}

\subsubsection{Overall Performance.}
Table~\ref{tab:overall} compares the overall performance with previous work. We see that, modern neural networks are considerably better than traditional KN models with or without templates. Moreover, our base model (with content-attention only) outperforms \newcite{wikibio}, showing our better engineering efforts. After adding up all proposed components, we obtain +2.5 BLEU and ROUGE improvement and +0.3 NIST improvement, achieving new state-of-the-art results.

\subsubsection{Ablation Test.}
Table~\ref{tab:ablation} provides an extensive ablation test to verify the effectiveness of each component in our model. The top half of the table shows the results without the copy mechanism, and the bottom half incorporates the copying score as described previously. We observe that the copy mechasnim is consistently effective with different types of attention.

We then compare content-based attention and link-based attention, as well as their hybrid (also Table~\ref{tab:ablation}). The results show that, link-based attention alone is not as effective as content-based attention. However, we achieve better  performance if combining them together with an adaptive gate, i.e., the proposed hybrid attention.
The results are consistent in both halves of Table~\ref{tab:ablation} (with or without copying) and in terms of all metrics (BLEU, ROUGE, and NIST). This implies that content-based attention and link-based attention do capture different aspects of information, and their hybrid is more suited to the task of table-to-text generation.

\subsubsection{Effect of the gate.}

\begin{figure}[!t]
	\centering
	\includegraphics[width=.7\linewidth]{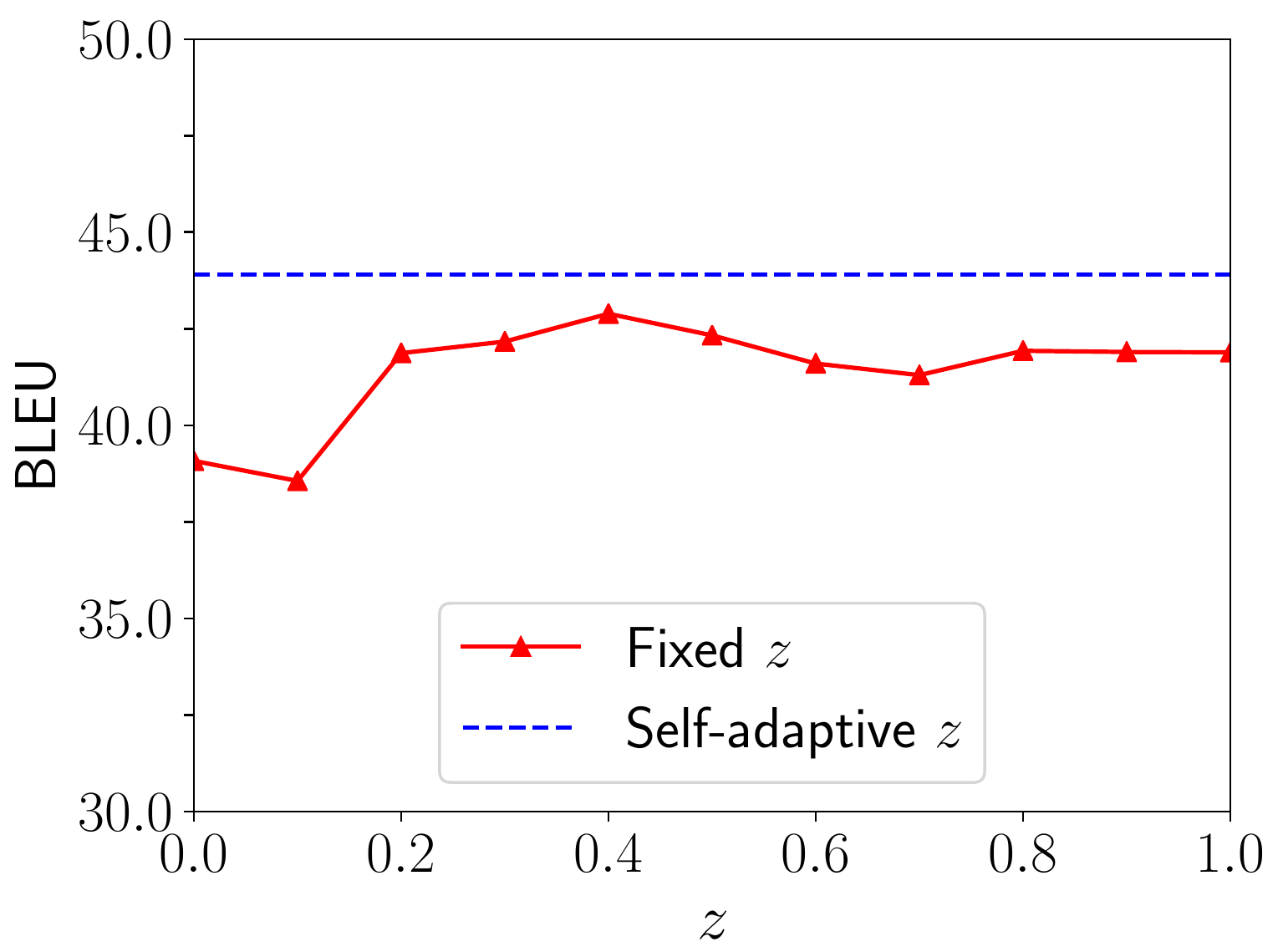}
	\vspace{-.3cm}
	\caption{Comparing the self-adaptive gate with interpolation of content- and link-based attention. $z=0$ is link-based attention, $z=1$ is content-based attention.}\label{fig:gate}
\end{figure}
We are further interested in the effect of the gate $z$, which balances content-based attention $\bm\alpha^\text{content}$ and link-based attention $\bm\alpha^\text{link}$. As defined in Equation~\ref{eq:gate}, the computation of $z$ depends on the decoding state as well as table information; hence it is ``self-adaptive.'' We would like to verify if such adaptiveness is useful. To verify this, we designed a controlled experiment where the gate $z$ was manually assigned in advance and fixed during training. In other words, the setting was essentially a (fixed) interpolation between $\bm\alpha^\text{content}$ and $\bm\alpha^\text{link}$. Specifically, we tuned $z$ from $0$ to $1$ with a granularity of $0.1$, and plot  BLEU scores as the comparison metric in Figure~\ref{fig:gate}.

As seen, interpolation of content- and link-based attention is generally better than either  single mechanism, which again shows the effectiveness of hybrid attention. However, the peak performance of simple interpolation (42.89 BLEU when $z=0.4$) is worse than the self-adaptive gate, implying that our gating mechanism can automatically adjust the importance of $\bm\alpha^\text{content}$ and $\bm\alpha^\text{link}$ at a particular time based on the current state and input.

\subsubsection{Different Ways of Using Field Information.}
We are curious whether the proposed order-planning mechanism is better than other possible ways of using field information. We conducted two controlled experiments as follows. Similar to the proposed approach, we multiplied the attention probability by a field matrix and thus obtained a weighted field embedding. We fed it to either (1) the computation of content-based attention, i.e., Equations~\ref{eqn:content1}--\ref{eqn:content2}, or (2) the RNN decoder's input, i.e., Equation~\ref{eqn:RNNinput}. In both cases, the last step's weighted field embedding was concatenated with the embedding of the generated word $\bm y_{t-1}$.

From Table~\ref{tab:field}, we see that feeding field information to the computation of $\bm \alpha^\text{content}$ interferes content attention and leads to performance degradation, and that feeding it to decoder RNN slightly improves model performance. However, both controlled experiments are worse than the proposed method. The results confirm that our order-planning mechanism is indeed useful in modeling the order of fields, outperforming several other approaches that use the same field information in a na\"ive fashion.

\begin{table}[!t]
	\centering
	\begin{tabular}{lccc}
		\toprule
		\!\!\textbf{Feeding field info to\dots} & \textbf{BLEU} & \textbf{ROUGE} & \textbf{NIST}\\
		\midrule
		None   &41.89 &34.93 & 8.63\\
		Computation of $\bm \alpha^\text{content}$  &40.52&34.95&8.57\\
		Decoder RNN's input  & 41.96 & 35.07 & 8.61\\
		\midrule
		Hybrid att. (proposed) &\textbf{43.91} &\textbf{37.15} & \textbf{8.85}\\
		\bottomrule
	\end{tabular}
	\caption{Comparing different possible ways of using field information. ``None'': No field information is fed back to the network, i.e., content-based attention computed by Equation~\ref{eqn:content} (with copying).}
	\label{tab:field}
\end{table}

\subsection{Case Study and Visualization}

\begin{table*}
	\centering
	\begin{minipage}[c]{0.24\textwidth}
		\begin{center}
			\includegraphics[width=\textwidth]{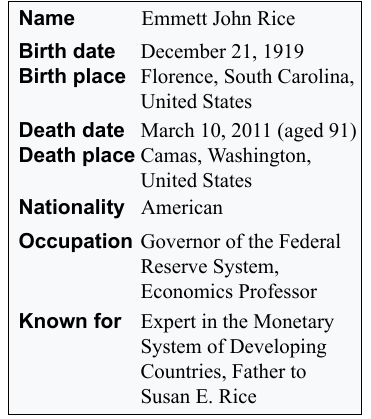}\\
		\end{center}
	\end{minipage}
	\begin{minipage}[c]{0.67\textwidth}
		\resizebox{\textwidth}{!}{
			\begin{tabular}{p{2.3cm}p{9.5cm}}\toprule
				Reference & emmett john rice ( december 21 , 1919 -- march 10 , 2011 ) was a former governor of the federal reserve system , a Cornell university economics professor , expert in the monetary systems of developing countries and the father of the current national security advisor to president barack obama , susan e . rice .\\\midrule
				Content-based attention& emmett john rice ( december 21 , 1919 -- march 10 , 2011 ) was an economist , author , public official   and the  former american governor of the federal reserve system , the first african american UNK .\\\midrule
				Hybrid attention&  emmett john rice ( december 21 , 1919 -- march 10 , 2011 ) was an american economist , author , public official  and the former governor of the federal reserve system , expert in the monetary systems of developing countries . \\
				\bottomrule
			\end{tabular}
		}
	\end{minipage}
	\vspace{-.1cm}	
	\caption{Case study. \textbf{Left}: Wikipedia infobox. \textbf{Right}: A reference and two generated sentences by different attention (both with the copy mechanism).}
    \vspace{-.4cm}	
	\label{tab:expsen}
\end{table*}

\begin{figure}[!t]
	\centering
	\includegraphics[width=\linewidth]{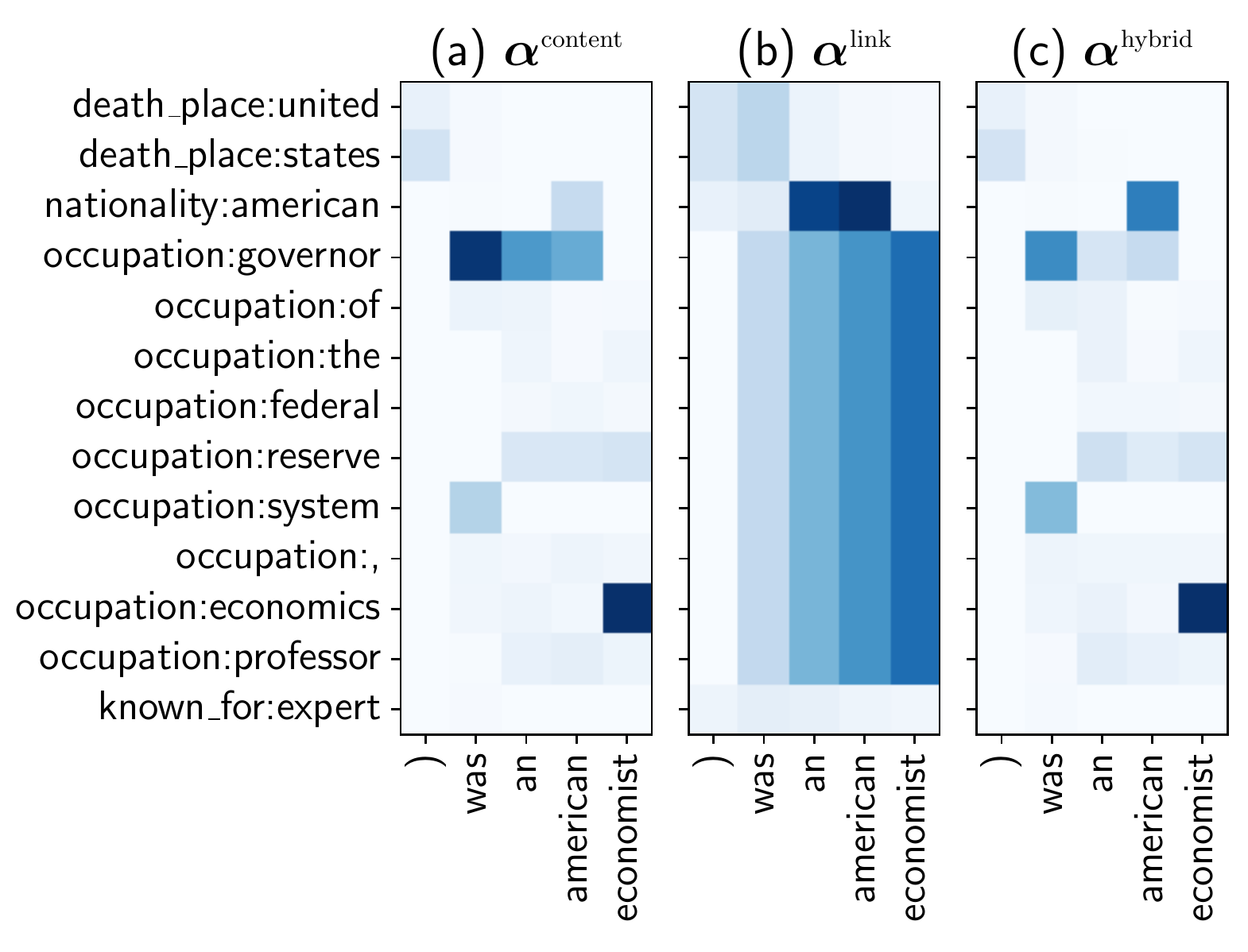}
	\vspace{-.7cm}
	\caption{Visualization of attention probabilities in our model. x-axis: generated words ``\dots ) was an american economist \dots ''; y-axis: $\langle$field : content word$\rangle$ pairs in the table. (a) Content-based attention.  (b) Link-based attention. (c) Hybrid attention. Subplot (b) exhibits strips because, by definition, link-based attention will yield the same score for all content words with the same field. Please also note that the columns do not sum to 1 in the figure because we only plot a part of the attention probabilities.}
	\vspace{-.2cm}
	\label{fig:vis}
\end{figure}

We showcase an example in Table~\ref{tab:expsen}. With only content-based attention, the network is confused about when the word \textit{American} is appropriate in the sentence, and corrupts the phrase \textit{former governor of the federal reserve system} as appears in the reference. However, when link-based attention is added, the network is more aware of the order between fields ``Nationality'' and ``Occupation,'' and generates the nationality \textit{American} before the occupation \textit{economist}. This process could also be visualized in Figure~\ref{fig:vis}. Here, we plot our model's content-based attention, link-based attention and their hybrid. (The content- and link-based attention probabilities may be different from those separately trained in the ablation test.) After generating ``\textit{emmett john rice ( december 21, 1919 -- march 10, 2011 ) was},'' content-based attention skips the nationality and focuses more on the occupation. Link-based attention, on the other hand, provides a strong clue suggesting to generate the nationality first and then occupation. In this way, the obtained sentence is more compliant with conventions.

\section{Related Work}

Text generation has long aroused interest in the NLP community due to is wide applications including automated navigation~\cite{navigate} and weather forecasting~\cite{weather}. Traditionally, text generation can be divided into several steps~\cite{NLG}: \textit{content planning} defines what information should be conveyed in the generated sentence; (2) \textit{sentence planning} determines what to generate in each sentence; and (3) \textit{surface realization} actually generates those sentences with words.

In early years, surface realization is often accomplished by templates~\cite{template} or statistically learned (shallow) models, e.g., probabilistic context-free grammar~\cite{pcfg} and language models~\cite{LM}, with hand-crafted features or rules. Therefore, these methods are weak in terms of the quality of generated sentences. For planning, researchers also apply (shallow) machine learning approaches. \newcite{collective}, for example, model it as a collective classification problem, whereas \newcite{semimarkov} use a generative semi-Markov model to align text segment and assigned meanings. Generally, planning and realization in the above work are separate and have difficulty in capturing the complexity of language due to the nature of shallow models.

Recently, the recurrent neural network (RNN) is playing a key role in natural language generating. As RNN can automatically capture highly complicated patterns during end-to-end training, it has successful applications including machine translation~\cite{attention}, dialog systems~\cite{dialog}, and text summarization~\cite{summarization}.

Researchers are then beginning to use RNN for text generation from structured data. \newcite{mei} propose a coarse-to-fine grained attention mechanism that selects one or more records (e.g., a piece of weather forecast) by a precomputed but fixed probability and then dynamically attends to relevant contents during decoding. \newcite{wikibio} incorporate the copy mechanism~\cite{copynet} into the generation process. However, the above approaches do not explicitly model the order of contents. It is also nontrivial to combine traditional planning techniques to such end-to-end learned RNN.

Our paper proposes an order-planning approach by designing a hybrid of content- and link-based attention. The model is inspired by hybrid content- and location-based addressing in the Differentiable Neural Computer~\cite[DNC]{DNC}, where the location-based addressing is defined heuristically. Instead, we propose a transition-like link matrix that models how likely a field is mentioned after another, which is more suited in our scenario.

Moreover, our entire model is differentiable, and thus the \textit{planning} and \textit{realization} steps in traditional language generation can be learned end-to-end in our model.

\section{Conclusion and Future Work}

In this paper, we propose an order-planning neural network that generates texts from a table (Wikipedia infobox). The text generation process is built upon an RNN with attention to table contents. Different from traditional content-based attention, we explicitly model the order of contents by a link matrix, based on which we compute a link-based attention. Then a self-adaptive gate balances the content- and link-based attention mechanisms. We further incorporate a copy mechanism to our model to cope with rare or unseen words.

We evaluated our approach on a newly proposed large scale dataset, \textsc{WikiBio}. Experimental results show that we outperform previous results by a large margin in terms of BLEU, ROUGE, and NIST scores. We also had extensive ablation test showing the effectiveness of the copy mechanism, as well as the hybrid attention of content and linking information. We compared our order-planning mechanism with other possible ways of modeling field; the results confirm that the proposed method is better than feeding field embedding to the network in a na\"ive fashion.
Finally we provide a case study and visualize the attention scores so as to better understand our model.

In future work, we would like to deal with text generation from multiple tables. In particular, we would design hierarchical attention mechanisms that can first select a table containing the information and then select a field for generation, which would improve the attention efficiency. We would also like to apply the proposed method to text generation from other structured data, e.g., a knowledge graph.

\section{Acknowledgments}
We thank Jing He from AdeptMind.ai for helpful discussions on different ways of using field information. 
\bibliographystyle{aaai}
\bibliography{refsaaai}

\end{document}